\documentclass{article} 
\usepackage{iclr2019_conference,times}

\usepackage{amsmath,amsfonts,bm}

\def\eqref#1{equation~\ref{#1}}

\def\1{\bm{1}}

\def\mI{{\bm{I}}}

\DeclareMathAlphabet{\mathsfit}{\encodingdefault}{\sfdefault}{m}{sl}
\SetMathAlphabet{\mathsfit}{bold}{\encodingdefault}{\sfdefault}{bx}{n}

\DeclareMathOperator*{\argmax}{arg\,max}

\usepackage{hyperref}
\usepackage{url}

\usepackage{natbib}
\usepackage{amsmath,amssymb,amsthm,bm}
\usepackage{algorithmicx,algorithm}
\usepackage{algpseudocode}
\usepackage{graphicx,subcaption,wrapfig}

\algblockdefx{MRepeat}{EndRepeat}{\textbf{repeat}}{}
\algnotext{EndRepeat}

\newtheorem{definition}{Definition}
\newtheorem{assumption}{Assumption}
\newtheorem{lemma}{Lemma}
\newtheorem*{remark*}{Remark}

\newcommand{\fan}{}

\title{Safety-Guided Deep Reinforcement Learning \\ via Online Gaussian Process Estimation}

\author{Jiameng Fan and Wenchao Li\\
Department of Electrical and Computer Engineering\\
Boston University\\
Boston, MA 02215, USA \\
\texttt{\{jmfan, wenchao\}@bu.edu} \\
}

\iclrfinalcopy 
\begin{document}

\maketitle

\begin{abstract}
	An important facet of reinforcement learning (RL) has to do with how the agent goes about exploring the environment.
Traditional exploration strategies typically focus on efficiency and ignore safety. 
However, for practical applications, ensuring safety of the agent during exploration is crucial since performing an unsafe action or reaching an unsafe state could result in irreversible damage to the agent.
The main challenge of safe exploration is that characterizing the unsafe states and actions is difficult for large continuous state or action spaces and unknown environments.
In this paper, we propose a novel approach to incorporate estimations of safety to guide exploration and policy search in deep reinforcement learning.  
By using a cost function to capture trajectory-based safety, our key idea is to formulate the state-action value function of this safety cost as a candidate Lyapunov function and extend control-theoretic results to approximate its derivative using online Gaussian Process (GP) estimation.
We show how to use these statistical models to guide the agent in unknown environments to obtain high-performance control policies with provable stability certificates.
\end{abstract}

\section{Introduction}
\label{introduction}
Deep reinforcement learning (RL) algorithms~\citep{sutton2018reinforcement} have achieved impressive results in game environments such as those on the Atari platform~\citep{mnih2015human}. However, they are rarely applied to real-world, physical systems. The main reason is that, besides the goal of optimizing for performance, there often exist safety requirements that make RL challenging in actual applications. In particular, these safety requirements might be imposed in deployment~\citep{amodei2016concrete, garcia2015comprehensive} or during exploration and training~\citep{leike2017ai, berkenkamp2017safe, chow2018lyapunov}. For example, an intermediate, learned policy exercised by a robot during training should not break the system or harm the environment. The importance of safety is well recognized by the community and safe reinforcement learning has recently emerged as an important subfield within RL (for an extensive survey, see \citet{garcia2015comprehensive}). In general, the goal of safe RL is to maximize system performance while minimizing safety violations (or meeting safety constraints) during the learning and/or deployment processes.

In this paper, we consider a notion of safety that is defined over executions of the agent (i.e., trajectories).
It has been observed that, in many safety-critical applications such as robot exploration~\citep{moldovan2012safe}, portfolio planning~\citep{tamar2012finance} and resource allocation~\citep{tesauro2006hybrid}, it is often more natural to define safety over the whole trajectory, as opposed to over particular states or state-action pairs.
We associate a real-valued \textit{safety cost} with each state-action pair.
A policy is thus deemed safe if its cumulative safety costs (different from the reward return) for the length of the trajectory is below a certain threshold. 
In general, this threshold might not be known \textit{a priori}. Thus, our goal is to keep the cumulative safety cost as low as possible.
Compared with approaches that guarantee safety over state-action pairs by relying on human oversight and intervention~\citep{saunders2018trial} or blocking the unsafe actions using the so-called shields~\citep{alshiekh2017safe}, trajectory-based safety is more suitable for evaluating the safety of a given policy when the environment model is unknown. Besides, characterizing unsafe states and unsafe actions can be intractable or infeasible for the high-dimensional and continuous cases. \fan{\citet{achiam2017constrained} proposed a method called \emph{constrained policy optimization} (CPO) that considers similar trajectory-based constraints and solves the problem in the setting of Constrained Markov Decision Processes. Although this method has good scalability and obtains safe policy during training, it is non-trivial to generalize the same framework beyond policy-gradient-based methods and improve sample efficiency in on-policy settings.}

\begin{figure}[t]
    \center
    \includegraphics[width=0.6\linewidth]{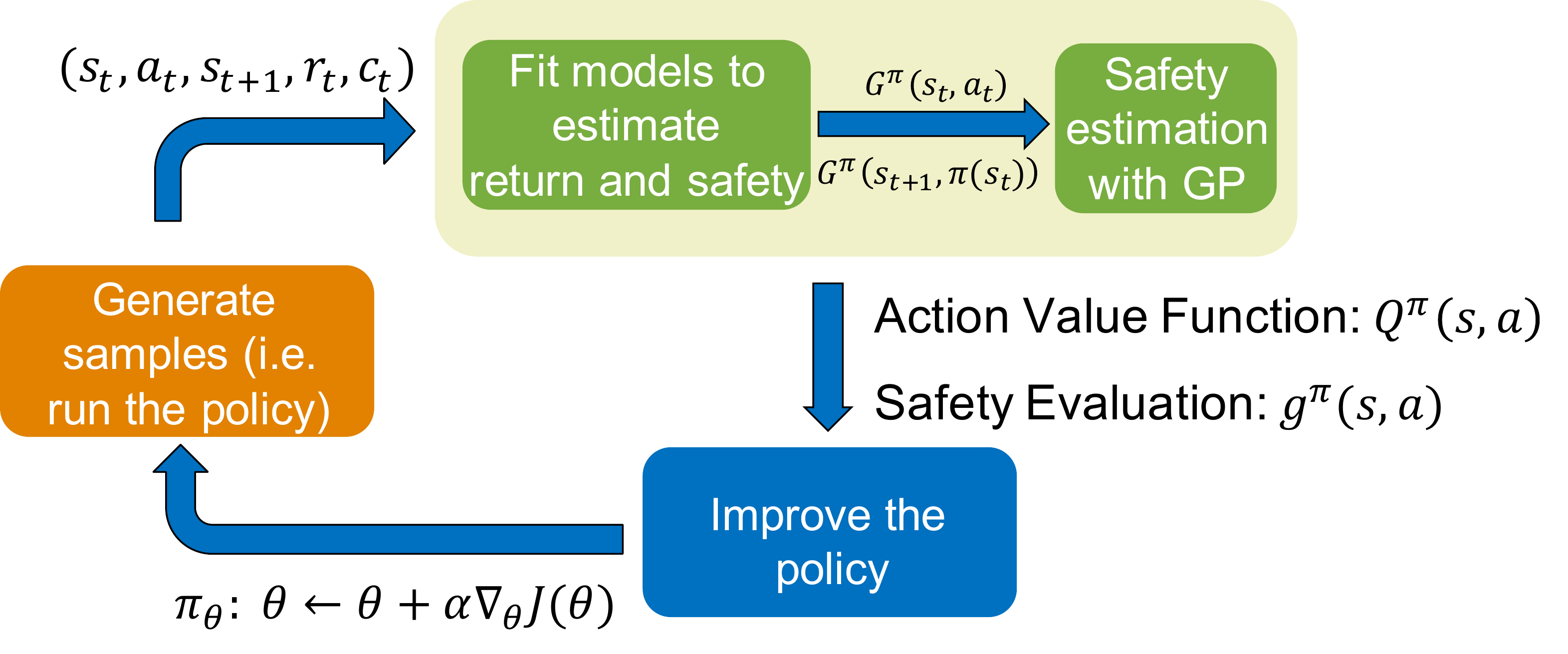}
    \caption{The safety-guided RL framework: the parameterized policy generates $(s_t, a_t, s_{t+1}, r_t, c_t)$ which includes current state, current action, next state, reward and safety cost along the trajectory; these values are used to fit models $Q^\pi(s,a)$ and $G^\pi(s,a)$ which estimate the expected reward and safety cost respectively; the GP estimation is updated in every iteration given the new tuples and measurements from $G^\pi(s,a)$; the parameterized policy is then optimized based on the objective function $J(\theta)$ which combines the reward return and safety estimations.}
    \label{fig:framework}
\end{figure}

In trajectory-based safety, in order to minimize the cumulative safety costs, it is important for the agent to be able to recover from states with high safety cost.
This ability to recover is known as \textit{asymptotic stability} in control theory~\citep{bhatia2002stability}, which provides a powerful paradigm to translate global properties of the system to local ones and vice versa. While the main challenge of Lyapunov-based methods~\citep{berkenkamp2016safe_CDC,bhatia2002stability} is to design an appropriate Lyapunov function candidate, our idea is to \textit{formulate the state-action value function for safety costs as the candidate Lyapunov function and model its derivative with a Gaussian Process which provides statistical guarantees}.
By combining with the original value function, our approach steers the policy search in a direction that both decreases the future cumulative safety costs and increases the expected total reward. 
Fig.~\ref{fig:framework} shows the overall framework.

In short, we propose \textit{a model-free RL algorithm that can provide high-probability trajectory-based safety guarantees for unknown environments with continuous state spaces}.
The main contributions of our paper are four-fold. 
\begin{itemize}
    \item We propose a novel Lyapunov-based approach to guide the exploration process of deep RL. 
    \item We propose to use Gaussian Processes to model the evolution of stability as policies get updated during training to cope with unknown environments and large continuous state/action spaces \fan{under off-policy settings}.
    \item We show that adjusting the GP estimation online is needed to effectively and safely guide policy search. 
    \item We demonstrate the effectiveness of the approach in significantly reducing the number of catastrophes (e.g. falling) during training and exploration in a high-dimensional locomotion task with continuous states and actions. In addition, we show that our approach can attain higher performance in fewer iterations and shorter amount of time compared to the Deep Deterministic Policy Gradient method.
\end{itemize}


\section{Related Work}
\label{related_work}
Safety is an important issue in RL and safe RL has emerged as an active research topic in recent years~\citep{pecka2014safe, garcia2015comprehensive}. Below, we discuss metrics of safety, representative approaches in model-based and model-free RL, and recent works on safe RL. 

\textbf{Safety Metrics.} The concept of safety, or dually, risk has taken various forms in the RL literature. 
In \citet{sato2001td}, the authors show that variability induced by the trained policy can lead to risky or undesirable situations. This characterization unfortunately does not generalize to settings where a policy with a small variance produces significant risks.
In general, the safety metric should be easily generalizable to any safety-critical domain and independent of the nature of the task. 
\citet{torrey2012help} propose a level metric based on the distance between the known and the unknown space. However, this metric relies on constant monitoring by humans to provide the necessary guidance. 
In \citet{gehring2013smart}, the authors measure safety as state controllability based on the notion of temporal difference.
The weighted sum of an entropy measurement and the expected return is used to evaluate safety in \citet{law2005risk}. 
While these metrics seem suitable for finite MDPs, for MDPs with large state and action spaces, these measurements are computationally intractable. This paper considers trajectory-based safety with respect to the executed policy and uses function approximators to estimate safety instead of relying on human monitoring or assuming that the MDP model is given.

\textbf{Model-based and Model-free RL.} In the model-based setting, research has focused on estimating the true model of the environment by interacting with it. Model-based methods typically cannot cope with continuous or large state/action spaces and have trouble scaling due to the curse of dimensionality~\citep{abbeel2005exploration}. 
In \textit{continuous} state/action spaces, model-free policy search algorithms have been shown to be successful. These approaches update the policies without knowing the system model by repeatedly executing the same task~\citep{lillicrap2015continuous}. 
\citet{achiam2017constrained} introduce safety guarantees in terms of constraint satisfaction that holds in expectation. However, safety has only been considered by disallowing large steps along the gradient into areas of the parameter space that have not been explored before. Existing works use
Gaussian Process models~\citep{rasmussen2004gaussian} along with Bayesian optimization~\citep{mockus2012bayesian} to approximate the value function~\citep{chowdhary2014off}.
On the down side, these methods are limited to simple and low-dimensional systems.

\textbf{Safe RL.} There are primarily two types of approaches to the safe RL problem: approaches that modify the optimization criterion with a safety component, and approaches that modify the exploration process through the incorporation of external knowledge~\citep{garcia2015comprehensive}. 

In RL, maximizing the long-term reward does not necessarily avoid the rare occurrences of large negative outcomes. In risk-sensitive RL, the optimization criterion is transformed into an exponential utility function~\citep{howard1972risk}, or a linear combination of return and risk, where risk can be defined as the variance of the return~\citep{sato2001td}. \citet{geibel2005risk} define risk as the probability of driving the agent to a set of known but undesirable states. The optimization objective is then transformed to include minimizing the probability of visiting those states.

Other works instead change the exploration process directly. Most exploration methods are based on heuristics and have a random exploratory component, which can result in the exploration being risk-blind. Both \citet{moldovan2012safe} and \citet{berkenkamp2017safe} introduce algorithms to safely explore state-action space so that the agent never gets stuck. However, these two methods require an accurate probabilistic or approximated statistical model of the system. The common shortcoming of these methods is that they are limited to small and simple systems where exact control synthesis is possible. \citet{eysenbach2017leave} propose to learn both forward and reset policies simultaneously with two action-value functions using Deep RL. Although the reset policy can move the agent back to the initial state after early aborts, there are no performance guarantees for the reset policy and the switching mechanism may result in very conservative behavior of the agent.

It is worth noting that the first type of approach, which modifies the optimization objective, will also modify the exploration process indirectly~\citep{garcia2015comprehensive}. The vital component across these two types of approaches is transforming the optimization criterion or change the exploration process to include a form of risk. In this paper, we propose a novel risk/safety evaluation-guided training technique that significantly improves safety during training and exploration.


\section{Background}
\label{background}
We consider a model-free RL setup, where an agent interacts with the environment $E$ in discrete timesteps.
RL is a sequential decision problem with state space $\mathcal{S}$, action space $\mathcal{A}$, transition dynamics $P(s'|s,a)$, an initial state distribution $p_0(s)$, and an immediate scalar reward $r(s,a)$. We need to specify a deterministic policy $\pi: \mathcal{S} \rightarrow \mathcal{A}$, that given the current state, determines the appropriate action that maximizes the expected sum of $\gamma$-discounted returns, $\mathbb{E} \left [ \sum_{t=0}^{T} \gamma^t r(s_t, a_t) \right ]$.

Typically, the RL training routines involve iteratively sampling from the current policy to explore the state-action space without considering safety. As a result, in practical applications, hard-coded termination or human intervention is required to stop the agent from entering unsafe states. Our work aims to enable safe exploration even when the environment is unknown or only partially known to us. Similar to the notion of reward, we define an additional function $c(s, a) \in \mathbb{R}_{\le0}$ as the negation of \textit{safety cost} to capture the cost of performing action $a$ in state $s$ with respect to safety. In the trajectory-based setting, the agent should aim to minimize future accumulated safety costs in a way similar to maximizing expected return. 
Safety requirement is defined over the whole trajectory.
This mean that, during training, the agent will try to avoid increasing the total safety costs, and pick exploratory actions that can drive the system away from the trajectories that violate the safety requirement.

\textbf{Deep Deterministic Policy Gradient (DDPG).} \citet{lillicrap2015continuous} proposed a model-free algorithm for solving the deterministic policy gradient problems with continuous action space. 
Let $\pi$ represent the deterministic policy.
\fan{Since the expectation depends only on the environment, it is possible to learn a state-action value function, $Q^\pi(s,a){=}\mathbb{E}_{s_{t+1} {\sim} E} \left [ r(s_t,a_t) {+} Q^{\pi}(s_{t+1}, \pi(s_{t+1})) \right ]$, off policy using transitions generated from another policy $\beta$ with different stochastic behaviors. Let $\rho^\beta$ be the state visiting distribution generated from $\beta$. 
DDPG combines greedy policy $\mu(s) = \argmax_a Q^\pi(s,a)$, which is commonly used in Q-learning~\citep{watkins1992q}, with function approximator $Q(s,a)$ and policies parameterized by $\theta^Q$ and $\theta^\pi$ respectively under the actor-critic architecture.}

Then, we can compute the gradient of the greedy policy by applying the chain rule to the expected return $J$ from the start distribution with respect to the actor parameters~\citep{lillicrap2015continuous}:

\begin{equation}
\nabla_{\theta^\pi} J= \mathbb{E}_{s_t \sim \rho^\beta} \left [ \nabla_a Q(s, a | \theta^Q)|_{s=s_t, a=\pi(s_t)} \nabla_{\theta^\pi}\pi(s|\theta^\pi)|_{s=s_t} \right ]
\end{equation}

\textbf{Lyapunov function.} To satisfy the specified safety requirement for safe exploration, we need a tool to determine safety of a trajectory that follows the current policy into the future. In control theory, this safety is usually computed for a fixed policy using Lyapunov functions. 
\begin{definition}
Lyapunov functions are continuously differentiable functions $V: \mathcal{X} \rightarrow \mathbb{R}_{\le 0}$ with $V(x) = 0, \forall x \in \mathcal{S}_0$ and $V(x)<0, \forall x \in \mathcal{X} \backslash \mathcal{S}_0$. The origin set $\mathcal{S}_0$ is set as the set of terminal states.
\end{definition}
\fan{In our algorithm, we leverage the fact that if the cost function $c(s,a)$ is strictly negative away from the origin and equal to zero at the origin, the action-value function of the accumulated costs, $G^\pi(s,a)$, in RL are Lyapunov functions.} This follows directly from the definition of the action-value function, where
\begin{equation}
G^{\pi}(s_t,a_t) = \mathbb{E}_{s_{t+1} \sim E} \left [ c(s_t,a_t) + G^{\pi}(s_{t+1}, \pi(s_{t+1})) \right ]
\end{equation}
\fan{We approximate $G^\pi(s,a)$ with an approximator $G(s,a)$ which is parameterized by $\theta_G$.}

\textbf{Safety Evaluation} The key idea is to use the Lyapunov function to provide the measurements of the trajectory-based safety. In recent literatures, trajectory-based properties are evaluated on a set of policies~\citep{achiam2017constrained, chow2018lyapunov}, which will require the function to be able to express the evaluation given some policy on the state-action space. Thus, we design the Lyapunov function as the accumulated safety costs $G^{\pi}(s,a)$ of policy $\pi$ with respect to $c(s,a)$.

We show that the state-action value function of safety cost is similar to that of gradient ascent on strictly quasiconcave functions: if one can show that, given a policy $\pi$, the agent is able to obtain strictly larger values of $G^\pi(s_t,a_t)$ at $t{+}1$ (‘going uphill’), then the state will eventually converge to the equilibrium points at the origin. Then, we can achieve safe exploration if $G^{\pi}(s_{t+1}, \pi(s_{t+1})) {-} G^{\pi}(s_t, a_t) {\ge} 0$ for given policy $\pi$. 
\fan{However, the model is not known a priori and only an approximation of $G^\pi$ can be obtained. Our idea is to use a Gaussian Process (GP) to model $g^\pi$, the difference between the outputs of $G^\pi$ in two consecutive timesteps along the system evolution given the current state-action pair.} Formally, during the training phase, the GP model, $g^\pi(s,a) {\sim} \mathcal{GP}(0, k((s,a), (s',a'))$, will be fed with approximated measurement $G(s_{t+1}, \pi(s_{t+1})) {-} G(s_t, a_t)$ at $(s_t ,a_t)$. In order to bound the safety evaluation, we make the following assumption.
\begin{assumption}
The function $g^\pi$ has bounded Reproducing kernel Hilbert space (RKHS) norm with respect to a continuously differentiable, bounded kernel $k(x, x')$; that is, $\Vert g^\pi \Vert_k \le B_g$.
\label{assump: RKHS_bound}
\end{assumption}
\fan{\begin{assumption}
We assume the valid approximated measurements $G(s_{t+1}, \pi(s_{t+1})) {-} G(s_t, a_t)$ are only corrupted by $\sigma$-sub-Gaussian noise (e.g. bounded in $[-\sigma, \sigma]$). In our case, the valid measurements should lie in the $\sigma$ ball of $c_t$ or $-c_t$. The value of $\sigma$ will be chosen according to the range of $c(s,a)$.
\label{assump: noise_bound}
\end{assumption}}


\section{Safe Exploration with GP Guidance}
\label{methodology}
We choose DDPG~\citep{lillicrap2015continuous} as the baseline RL algorithm, since its off-policy learning allows sharing of the experience between the expected return of reward and safety costs estimation.

\subsection{Approximate Lyapunov Function}
\fan{We consider an additional function approximator, namely the Guard Network $G$, parameterized by $\theta^G$ to approximate $G^\pi$, that minimizes the following loss.}
\begin{equation}
L(\theta^G) = \mathbb{E}_{s_t \sim \rho^\beta} \left [ G(s_t, a_t\: |\: \theta^G) - y_t^2 \right ]
\end{equation}
\begin{equation}
\text{where} \quad y_t = c(s_t, a_t) + G(s_{t+1}, \pi(s_{t+1})\:|\:\theta^G)
\end{equation}

\subsection{Gaussian Process}
\fan{In GP regression, we use the Guard Network to compute the $G^\pi$ difference between two consecutive timesteps as noisy observations of the true safety estimation.}
Let $z = (s, a)$ denote the station-action pair observed by GP. Specifically, we can obtain the posterior distribution of a function value $g^\pi(z)$ at arbitrary state-action pair by conditioning the GP distribution of $g^\pi$ on a set of past measurements with $\sigma$-bound noise, $y_n=\{ \hat{g}^\pi(z_1), \dots, \hat{g}^\pi(z_n) \}$ for state-action pairs $\mathcal{D}_n = \{z_1, \dots, z_n\}$. The measurements are provided by the Guard Network approximation given the current policy, current state-action pair and the next state:
\begin{equation}
\hat{g}^\pi(s_t, a_t) = G(s_{t+1}, \pi(s_{t+1})\:|\:\theta^G) - G(s_t, a_t\:|\:\theta^G)
\label{eq:bound}
\end{equation}
\fan{To collect the valid observations, we select the measurements within the $\sigma$ balls of $c_t$ or $-c_t$.} The posterior over $g^\pi(z)$ is a GP distribution again, with mean $\mu_n(z)$, covariance $k_n(z, z')$ and variance $\sigma_n(z)$.
\begin{align}
\mu_n(z) &= k_n(z)(K_n+\displaystyle \mI_n \sigma^2)^{-1}y_n \label{eq:mean} \\
k_n(z, z') &= k(z, z') - k_n(z)(K_n+\displaystyle \mI_n \sigma^2)^{-1}k_n^T(z') \label{eq:convariance} \\
\sigma_n^2(z) &= k_n(z, z)
\label{eq:variance}
\end{align} 
where $k_n(z) = (k(z, z_1), \dots, k(z, z_n))$ contains the covariances between the new input $z$ and $z_i$ in $\mathcal{D}_n$, $K_n \in \mathbb{R}^{n \times n}$ is the positive-definite covariance matrix. $\displaystyle \mI_n \in \mathbb{R}^{n \times n}$ is the identity matrix.

With Assumption~\ref{assump: RKHS_bound} and~\ref{assump: noise_bound} we can obtain the following result for $g^\pi(z)$~\citep{chowdhury2017kernelized}:
\begin{lemma}
Supposed that $\Vert g^\pi\Vert_k^2 \le B_g$ and that the observation noise is uniformly bounded by $\sigma$. Choose $\beta_n = B_g^{1/2} + 4\sigma(\gamma_{n-1} + 1 + \ln(2/\delta))^{1/2} $, where $\gamma_n$ is the information capacity. Then, for all $n \ge 1$, it holds with probability at least $1-\delta$, $\delta \in (0,1)$ that
\begin{equation}
| g^\pi(z) - \mu_{n-1}(z) | \le \beta_n \sigma_n(z)
\end{equation}
\label{lemma: bound}
\end{lemma}
Lemma~\ref{lemma: bound} allows us to make high-probability statements about the true function values of $g^\pi(z)$. 
The information capacity, $\gamma_n = \max_{z_1, \dots, z_n} I(g^\pi, y_n)$, is the maximal mutual information that can be obtained about the GP prior from $n$ noisy samples $y_n$ at state-action pairs set $\mathcal{D}_n$. 
\fan{This function was shown to be sublinear in $n$ for many commonly-used kernels in~\cite{srinivas2009gaussian}. Details about the computation of this function can be found in the Appendix.}
As a result, we are able to learn about the true values of $g^\pi(z)$ over time by making appropriate choices from $\mathcal{D}_n$.

\subsection{Initialization}
To prevent our model from converging too quickly to an incorrect estimate of $G^\pi$ in high-dimensional tasks, we introduce a single safe trajectory, $\xi_{init}$ with state-action pairs at each timestep, as initial knowledge to initialize the GP model, the $Q$ approximator and the $G$ approximator. This trajectory is required to be safe in the sense that the cost measurements in each state are less than some threshold depending on the system requirement. \fan{Hence, we will only keep the state-action pairs that satisfy the cost threshold, which will not require a completely safe trajectory.}
These safe state-action pairs will be added to the replay buffers of the $Q$ and $G$ approximators with the associated rewards given by $r(s,a)$. The initial GP dataset $\mathcal{D}$ will contain these state-action pairs, and the measurements are given by the negation of cost function for each state-action pair as $-c(s,a)$. \fan{For low-dimensional tasks, we typically do not need to use such initial knowledge since the kernels in our GP model are less sensitive to low-dimensional inputs.}

\subsection{Online GP Estimation}
In order to incorporate new data, we maximize the marginal likelihood of $g^{\pi}(z)$ after every iteration by adjusting the hyperparameters of the GP model. The term marginal likelihood refers to the marginalization over the function values $g^\pi$. Under the Gaussian Process model, the prior is Gaussian, i.e. $g^\pi|\mathcal{D}_n \sim \mathcal{N}(0, K_n)$, and the likelihood is a factorized Gaussian, i.e. $y_n|g^\pi \sim \mathcal{N}(g^\pi, \sigma_n^2 I)$. We can then obtain the log marginal likelihood as follows~\citep{rasmussen2004gaussian}.

\begin{equation}
\log p(y_n|\mathcal{D}_n) = -\frac{1}{2} y_n^T (K_n + \sigma_n^2 \displaystyle \mI_n)^{-1} y_n - \frac{1}{2} \log |K_n + \sigma_n^2 \displaystyle \mI_n| - \frac{n}{2} \log 2\pi
\end{equation}
The hyperparameters in the GP model, such as the kernel function's parameters, can be optimized to fit the current dataset $\mathcal{D}$ and measurements $y_n$ with high probabilities. This step is aimed at addressing the issue of inaccuracy in the initial $G^\pi(s,a)$ estimation.

As an agent continues to collect new measurements during the execution of policies, the set of samples will increase in size. The state-action pair will be stored in $\mathcal{D}_n$ if the measurements, $\hat{g}^\pi(s,a)$, are outside the $\sigma$ ball of zero and are valid. \fan{We use this to prevent overfitting at the origin sets, which can result in very conservative (though safe) behaviors.} After each run, the singularity of the covariance matrix based on $\mathcal{D}_n$ will be checked by QR decomposition to eliminate highly correlated data.

\fan{Performing the prediction by computing Eq.~\ref{eq:mean} and Eq.~\ref{eq:convariance} requires an expensive inversion of a matrix that scales cubically with the size of the data, which means maintaining a large dataset is not practical.}
If we maintain a dataset of fixed size, a natural and simple way to determine whether to delete a point from the dataset is to check how well it is approximated by the rest of the elements in $\mathcal{D}_n$. This is known as the kernel linear independence test~\citep{csato2002sparse}. For GPs, the linear independence test for the $i^{\text{th}}$ element from $\mathcal{D}_{n+1}$ is computed as
\begin{equation}
\phi(z_i) = k(z_i,z_i) - k_n(z_i)K_n^{-1}k_n^T(z_i)
\end{equation}
which is the variance of $z_i$ conditioned on the rest of $n$ elements without observation noise. In \citet{csato2002sparse}, they show that the diagonal values of $K_{n+1}^{-1}$ correspond to $\phi(z_i)$ of the $i^{\text{th}}$ element. Hence, we can delete the element that has the lowest value of $\phi$ such that it will have less impact on the GP prediction and keep the size of the dataset at $n$.
\begin{remark*}
While the full dataset $\mathcal{D}_{n}$ encounters a new data point and becomes $\mathcal{D}_{n+1}$, the kernel linear independence test will measure the length of the each data basis vector, $\tau_i$, in kernel space that is perpendicular to the linear subspace spanned by the current bases. For GPs, the linear dependence values vector $\tau$ for each data element in $\mathcal{D}_{n+1}$ can be computed as $\text{diag}(K_{n+1}^{-1})$.
\label{remark:online_gp}
\end{remark*}
\fan{Notice that the bound provided in Lemma~\ref{lemma: bound} only depends on the collected data in the current dataset. This means the online updates of GP can still provide the high-probability guarantees about the $g^\pi$ approximation.}

\subsection{Safety-Guided Exploration}
Given the result of Lemma 1, we can derive the lower and upper bounds of the confidence intervals after $(n-1)$ measurements of $g^\pi(s,a)$ from Eq.~\ref{eq:bound}
\begin{equation}
l_n(s,a) := \mu_{n-1}(s,a) - \beta_n \sigma_{n-1}
\label{eq:lower_bound}
\end{equation}
\begin{equation}
u_n(s,a) := \mu_{n-1}(s,a) + \beta_n \sigma_{n-1}
\label{eq:upper_bound}
\end{equation}
respectively. In the following, we assume that $\beta_n$ is chosen according to Lemma~\ref{lemma: bound}, which allows us to state that $g^\pi(s,a)$ takes values within $[l_n (s,a), u_n (s,a)]$ with high probability (at least $1 -\delta$).

Given the confidence interval, we can adapt our policy search to maximize the Q-value, while ensuring that the lower bound of $g_\pi(s, a)$, also the worst-case increase of Lyapunov function, is larger than zero with high probability. Thus, we construct the following constrained optimization problem:
\begin{equation}
\begin{aligned}
\pi(s) &= \argmax_a Q(s,a) \\
& \text{s.t. } l_n(s,a) \ge 0
\end{aligned}
\end{equation}
\fan{However, since in Eq.~\ref{eq:upper_bound} the lower bound is computed from GP prediction and the data-dependent parameter $\beta_n$, this constrained optimization problem cannot be solved directly
Instead, we softly enforce the safety requirement by picking a positive scalar $M$ and reformulate it as an unconstrained optimization problem as the following:
\begin{equation}
\pi(s) = \argmax_a Q(s,a) + M \cdot l_n(s,a)
\end{equation}
where $M$ is large enough to force the agent to choose the safe action satisfying $l_n(s,a) \ge 0$.}

\fan{To improve the accuracy of GP prediction, the exploration should not only satisfy the safety requirements but also reduce the uncertainty of the GP.} Thus, we select the policy in the following way.
\begin{align}
&\pi(s) = \argmax_a \sigma_{n-1}(s,a) \\
\text{s.t.}\quad &\mu_{n-1}(s,a) - \beta_n \sigma_{n-1} \ge 0
\end{align}
These two objectives will turn the safe exploration problem into a multi-objective optimization problem. On one hand, the agent will take a safe action to maximize the return. On the other, the chosen action should provide as much information as possible to the GP estimation to reduce uncertainties. From the above formulation, we can derive that the optimal value of action $a^*$ with the following property.
\begin{equation}
\mu_{n-1}(s,a^*) \ge \beta_n\sigma_{n_1}(s, a^*) \ge 0
\end{equation}

\begin{algorithm}[t]
\scriptsize
\caption{\small Safety-Guided DDPG}
\begin{algorithmic}
\State Initialization
\begin{itemize}
	\item Initialize GP data set $\mathcal{D}$, the replay buffers of Q and G with initial knowledge $\xi_{init}$ if state-action space is high-dimensional.
	\item Initialize GP model $g^\pi(s,a) \sim \mathcal{GP}(0, k((s,a), (s',a'))$ and the bound $\sigma$ on observation noise. 
\end{itemize}
\MRepeat
\State $\mathcal{D}_{temp} = \emptyset$
\For{$t=0$ to $T$}
\State $(s_{t+1}, r_t, c_t) \leftarrow $ Environment.step($\pi_{\theta^\pi}(s_t)$)
\State $y_t \leftarrow G(s_{t+1}, \pi_{\theta^\pi}(s_{t+1})\:|\:\theta^G) - G(s_t, a_t\:|\:\theta^G)$
\If{$y_t$ is in $\sigma$ ball of $c_t$ or $-c_t$}
	\State Store data element $((s_t, a_t), y_t)$ in $\mathcal{D}_{temp}$
\EndIf
\EndFor
\State Update $Q$ and $G$.
\State Concatenate $\mathcal{D}_{temp}$ with $\mathcal{D}$.
\While{D.size $>$ N}
\State Pick the first $N+1$ elements and remove the element with the lowest score, where scores = $\text{diag}(K_{N+1}^{-1})$.
\EndWhile
\State Update the actor policy $\pi_{\theta^\pi}$ via SGD on Eq.~\ref{eq:objective}
\EndRepeat
\end{algorithmic}
\label{alg:algorithm}
\end{algorithm}

With this property, we can combine these two objectives and constraints, with a term that penalizes the actions that result in negative lower bounds and rewards the actions that result in positive lower bounds around zero. Thus, we can design the term as a Gaussian distribution with zero mean for $l_n$. We can rewrite the multi-objective policy optimization problem using the weighted-sum method:
\begin{equation}
\pi(s) = \argmax_a Q(s,a) - M \cdot \rho(- l_n(s,a)) + exp(-l_n(s,a)^2)
\label{eq:objective}
\end{equation}
where $\rho(x) = max(0,x)$. So far, we have three components in the policy optimization objective, maximizing the reward return as given by the $Q$-value, penalizing violation of safety, and reducing uncertainty of GP.
The overall algorithm is summarized in Algorithm~\ref{alg:algorithm}.

\section{Experiments}
\label{experiments}
In this section, we evaluate Algorithm 1 on two different tasks in simulation, inverted pendulum and half cheetah from the OpenAI Gym~\citep{brockman2016openai}. We assume that the dynamics of the system and the environment are both unknown. We consider the performance of the trained vanila DDPG policy after $1$ million steps as the baseline.
We first validate our approach on a benchmark swing-up problem in the inverted pendulum environment. 
Then, we extend our experiment to a more complex and safety-critical locomotion task where the goal is to make a half cheetah move forward as fast as possible. Both environments are in continuous state/action space and initialized randomly for each run. The safety goal is that the number of catastrophes, as defined in each experiment, should be minimized during training.

\begin{wrapfigure}[21]{R}{0.48\linewidth}
    \vspace{-20pt}
    \center
    \includegraphics[width=\linewidth]{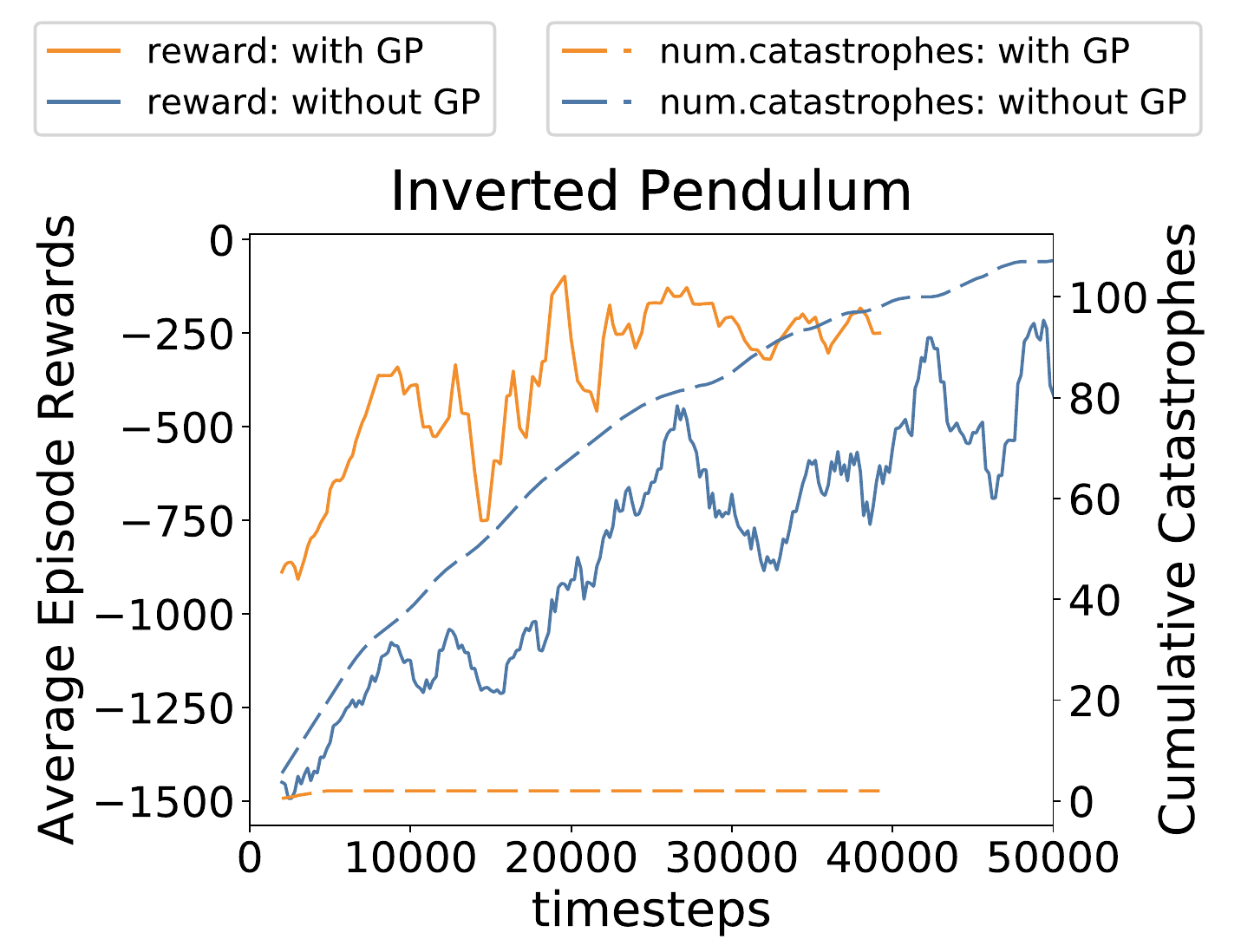}
    \caption{Comparison between DDPG with GP and the vanila DDPG on executing a swing-up task of an inverted pendulum. Both performance and the number of training-time catastrophes are plotted against timesteps. The average return achieved by DDPG after $500,000$ steps is $-244.9$.}
    \label{fig:pendulum_steps}
\end{wrapfigure}

\subsection{Inverted Pendulum}
The state of the inverted pendulum contains the angle $\theta$ and angular velocity $\dot{\theta}$ of the pendulum. The limited, applied torque is the action $a$.
The goal is to swing up and balance the pendulum in an upright position.
We define a negative reward which penalizes the large $\theta$, $\dot{\theta}$ and $a$.
In this case, the negation of the safety cost will be the same as the reward, which will lead the agent to swing up and stay at the vertically upward position.
We optimize the policy via stochastic gradient descent on Eq.~\ref{eq:objective}. More details about the settings are in Appendix~\ref{appendix: pendulum}

To improve the computation efficiency, we fix $\beta_n = 2$ in this experiment. In this case, catastrophe is defined as going through the vertically downward position in one episode (200 timesteps per episode). The experimental result~\footnote{\small Video link of the training result in pendulum environment: \url{https://youtu.be/etYqt15sGRY}} is shown in Fig.~\ref{fig:pendulum_steps}. Starting from a random initial state, the policy derived from DDPG with GP can avoid castastrophe entirely during training. The pendulum achieves the baseline performance after around $40,000$ steps, which is much less compared to the $500,000$ steps that vanila DDPG needs.

\subsection{Locomotion Task}
We further validate our approach on a 6-DOF planar half cheetah model with 17 continuous state components in MuJoCo~\citep{todorov2012mujoco}. Typically, in more complex tasks, it will be harder to encode both safety and performance in the same function. Also, the initial GP estimation will be very unreliable. Hence, we design different functions to represent reward and safety cost respectively, and assume some initial knowledge $\xi_{init}$ is given.
\begin{figure}[t]
    \centering
    \begin{subfigure}{0.6\linewidth}
    \centering
    \includegraphics[width=\linewidth]{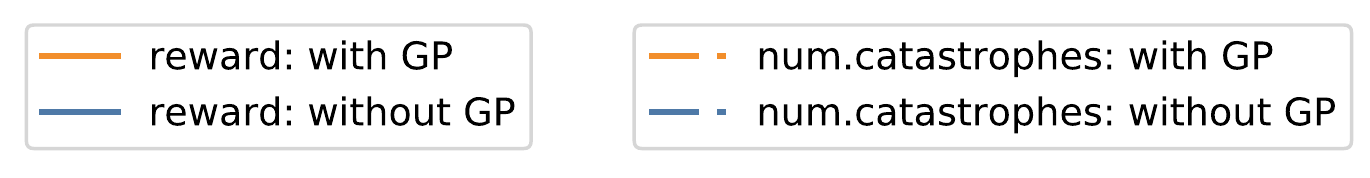}
    \end{subfigure}
    \vfill
    \begin{subfigure}{0.48\linewidth}
    \includegraphics[width=\linewidth]{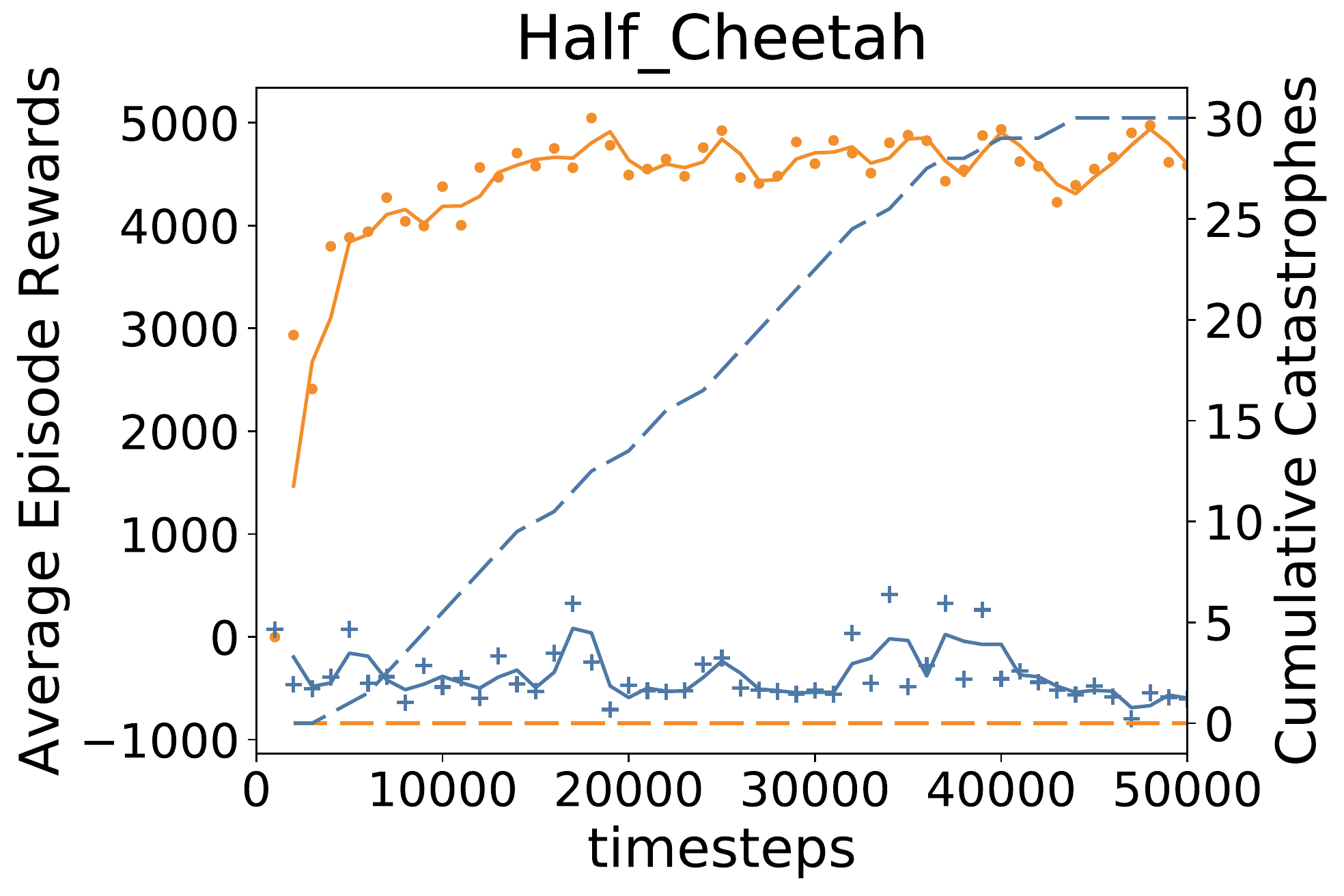}
    \caption{}
    \label{fig:results_cheetah_steps}
    \end{subfigure}
    \hfill
    \begin{subfigure}{0.48\linewidth}
    \includegraphics[width=\linewidth]{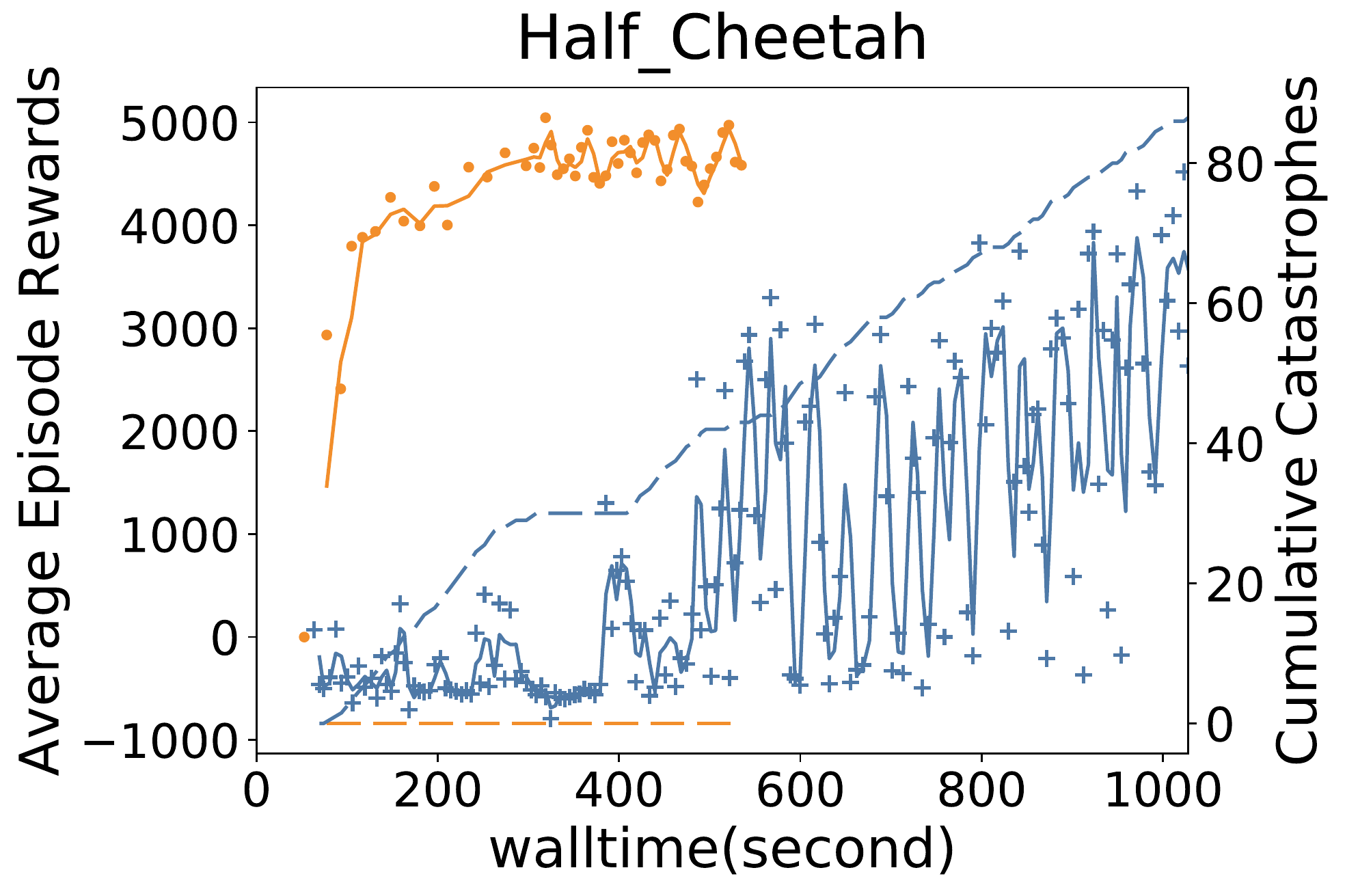}
    \caption{}
    \label{fig:results_cheetah_walltime}
    \end{subfigure}
    \caption{(a) The figure compares between DDPG with GP and vanila DDPG on half cheetah with the same initial knowledge. Performance and number of training-time catastrophes curves are by discrete timesteps. Vanila DDPG will achieve an average return of $4976.8$ only after $700,000$ steps.
    (b) Performance and number of training-time catastrophes are plotted against wall time.}
\end{figure}
\begin{wrapfigure}{R}{0.48\linewidth}
    \vspace{-10pt}
    \centering
    \begin{subfigure}{\linewidth}
    \centering
    \includegraphics[width=\linewidth]{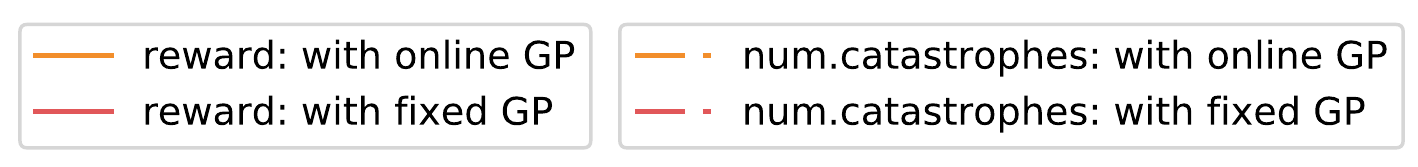}
    \end{subfigure}
    \vfill
    \begin{subfigure}{\linewidth}
    \includegraphics[width=\linewidth]{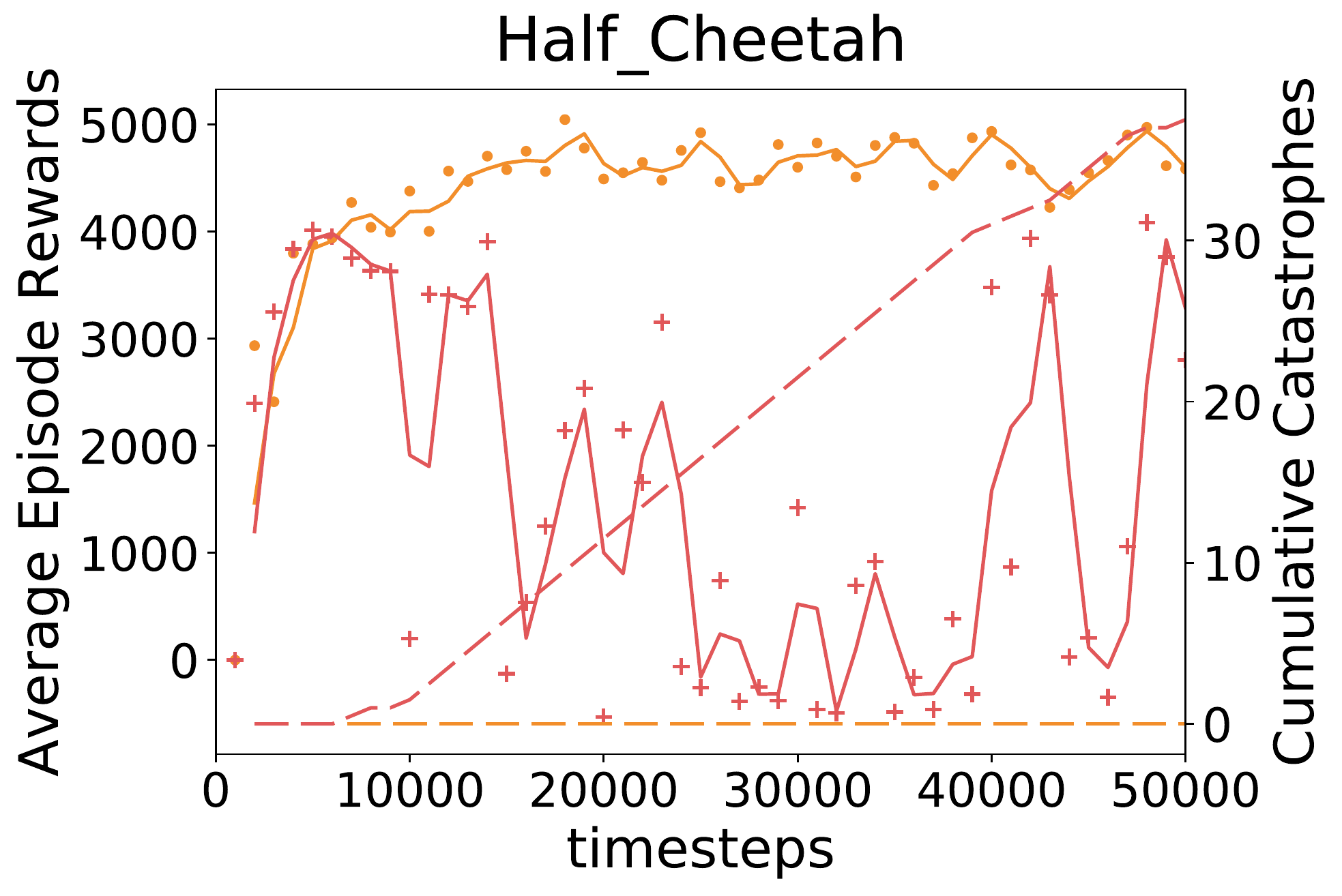}
    \end{subfigure}
    \caption{We compare DDGP with online GP and DDPG with fixed GP given the same initial knowledge. 
    Both have similar performance initially. As training progresses,
    online GP outperforms fixed GP significantly.}
    \label{fig:fixed_init}
\end{wrapfigure}


We design the reward function to maximize the forward speed and penalize control loss.
A catastrophe is considered to have occurred when the half cheetah falls down somewhere along the trajectory.
We cap the dataset for GP estimation to $2,000$ elements and initialize it with a single safe trajectory $\xi_{init}$ containing $1,000$ elements. The scaling factor for the confidence intervals, $\beta_n$, is approximated by the past $n-1$ measurements. More details about the settings are in Appendix~\ref{appendix: cheetah}

\textbf{Safety and Performance Comparison.}
For a fair comparison, we feed the same initial knowledge into the replay buffer of the vanila DDPG before training. Using our method, the agent can safely explore the environment and achieve the baseline performance after around $50,000$ steps as the vanila DDPG policy obtains after $700,000$ steps. We compare our method with vanila DDPG trained with the same amount of samples in Fig.~\ref{fig:results_cheetah_steps}. The result~\footnote{\small Video link of the training result in Cheetah environment: \url{https://youtu.be/CcNIrLlbijU}} shows our method obtains higher return and fewer training-time catastrophes than vanila DDPG. Although the prediction and data elimination from the online GP model will add computation overhead, DDPG with GP is still able to achieve higher performance and safer policy within the same amount of wall time (Fig.~\ref{fig:results_cheetah_walltime}).
Our approach is in line with recent results on learning acceleration when a small amount of demonstration data is available at the beginning~\citep{vevcerik2017leveraging, hester2018deep}.

\textbf{Validate the Role of Online GP.}
We compare safety-guided learning using online GP estimation with one that uses a fixed GP model. 
We initialize both models with the same initial knowledge. 
In Fig.~\ref{fig:fixed_init}, we can see that the initial performances of both models are similar. However, as training goes on, for DDGP with fixed GP, the accumulated reward drops and the number of training-time catastrophes increases (due to inaccuracies in the GP estimation). 
For the same number of timesteps, DDPG with fixed GP has lower performance than DDPG with online GP.
This result shows that adjusting the GP models online is critical as policies get updated during training.



\section{Conclusion} 
\label{conclusion}
In this paper, we propose to tackle the safe RL problem with the notion of Lyapunov function and trajectory-based safety to learn policies that are both safe and have low accumulated safety cost during exploration. We have shown how to incorporate estimation of trajectory-based safety in deep reinforcement learning algorithms such as DDPG. Specifically, we show how to safely optimize policies and give stability certificates based on Gaussian Process models of trajectory-based safety evaluation. On a simple control benchmark and a more complex locomotion task, we demonstrate the effectiveness of our approach in significantly reducing catastrophes and accelerating training.

In terms of future work, we want to understand better what role initial knowledge plays in influencing the efficacy of our method. One direction is to come up with statistical characterization of initial knowledge which can give statistical guarantees on the safety of the training process. On the computational side, as safety evaluation inevitably adds an overhead to the training process, we plan to investigate more efficient ways to estimate trajectory-based safety and to incorporate these estimates in policy optimization. 

{\small \bibliography{references}}
\bibliographystyle{iclr2019_conference}

\appendix
\section{Experiment Details}
\label{appendix}
\subsection{Experiment Settings}
For all of our examples, we represent the $Q$ function, $G$ function and the policy as three feed-forward neural networks with two hidden layers and variant neurons in the different environments. The settings is similar to \citet{lillicrap2015continuous}.

\subsection{Pendulum}
It has a single continuous action which is the applied torque bounded by $[-2, 2]$. The limited torque will make the task harder since the maximum applied torque will not be able to swing up the pendulum directly. 
We define the reward function $r(s,a) = s^T P s + a^T U a$, where the negative-definite $P$ and $R$ will penalize the large angular position $\theta$, angular velocity$\dot{\theta}$ and action $a$. The cost function is the same as the reward function, $c(s,a) = r(s,a)$.

To approximate the $Q$ function and $G$ function, we use a feed-forward neural network with two hidden layers, and each consists of 64 neurons. The hidden layers use the ReLU as the activation function, and the output layer does not use the activation function.
For the policy, we use a feed-forward neural network with two hidden layers and 64 neurons in each layer. We use ReLU for the hidden layers and tanh for the output layer.

\label{appendix: pendulum}

\subsection{Half-Cheetah}
The Half-Cheetah environment consists of 17 continuous states and 6 continuous action input each controls one of the six joints. We define a reward function $r(s,a) = v(s) - 0.1 \cdot a^Ta$ that rewards the positive forward velocity and penalizes the large control actions. The cost function here is related to the body rotation $\omega$, which is $c(s,a) = - \Vert \omega \Vert^2$. The larger value of $\omega$, the cheetah will be more likely to fall down, which is defined as catastrophes in this environment.

The $Q$ function and $G$ function are represented by two separated feed-forward neural networks with two hidden layers, and each consists of 64 neurons. The hidden layers use the ReLU as the activation function, and no activation function is applied at the output layers.
The policy network has $2$ hidden layers with $400$ and $300$ neurons respectively ($\approx 130,000$ parameters), which is the same used in \citet{lillicrap2015continuous}. The hidden layers implement with the ReLU function as the activation function and the output layer implement tanh function as the activation function.

Since in the high-dimensional space, it will be too conservative if we use a constant to approximate the scaling factor for the confidence intervals, $\beta_n$. Thus, we compute the approximated the scaling factor with the samples in the current dataset. The mutual information can be computed as:
\begin{equation}
    \gamma_{n-1} = \sum_{i=1}^{n-1}\log(1+\sigma_i((s,a)_i)/\sigma^2)
\end{equation}
and the RKHS bound can be obtained through kernel function as 
\begin{equation}
    B_g^2 = g_\pi((s,a))^T K_n  g_\pi((s,a))
\end{equation}
Thus, according to Lemma~\ref{lemma: bound}, we can compute $\beta_n$ online.
\label{appendix: cheetah}

\section{Extra Results}
\begin{figure}[h]
    \center
    \includegraphics[width=0.7\linewidth]{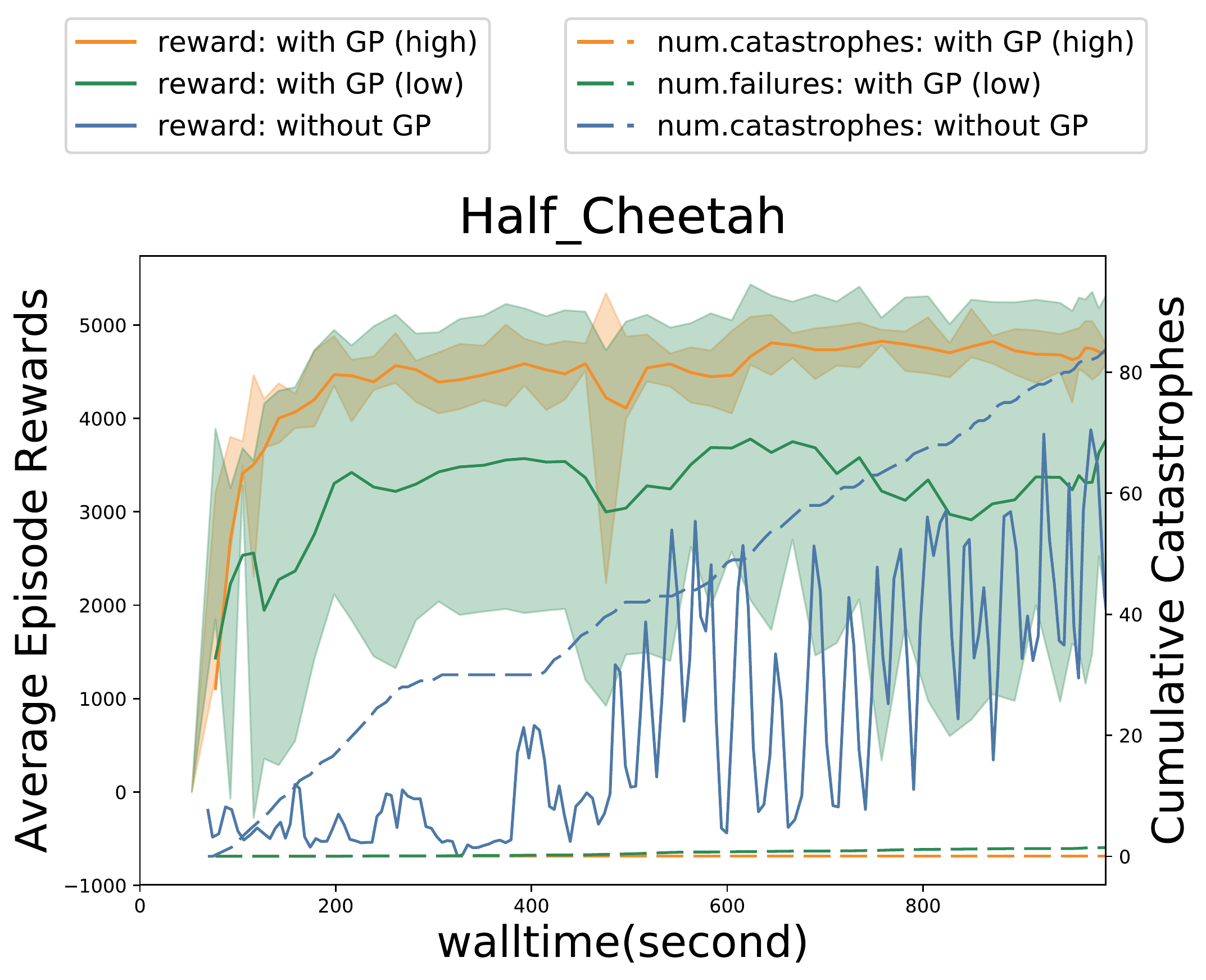}
    \caption{Comparison between DDPG with GP initialized by the high performance trajectory and the low performance trajectory for 7 runs.}
    \label{fig:init_comparison}
\end{figure}

In Fig.~\ref{fig:init_comparison}, we investigate the initial knowledge choices. Two full trajectories with different accumulated reward are considered here. The low performance trajectory obtains the return $2438.00$ and accumulated safety costs $64.95$. The high performance trajectory obtains the return $4985.20$ and accumulated safety costs $141.30$. The two initialization settings can both ensure the safety during the training. However, we can derive that the high performance trajectory is tend to guide the policy search more close to the optimal policy and results in less performance variance.

\end{document}